\documentclass[journal,twocolumn]{IEEEtran}

\usepackage{amsmath}
\usepackage{subfigure}
\usepackage{caption}
\usepackage{algorithm}
\usepackage{algpseudocode}
\usepackage{setspace}
\usepackage{color,hyperref}

\usepackage{graphicx}
\usepackage{amsmath,amssymb} 
\usepackage{epsfig}
\usepackage{amsmath}
\usepackage{amssymb}
\usepackage{colortbl}
\usepackage{soul}
\usepackage{multirow}
\usepackage{pifont}
\usepackage{bm}

\newcommand{\rmark}{\ding{52}}

\definecolor{mygray}{gray}{.9}

\newcolumntype{I}{!{\vrule width 1.2pt}}

\newlength\savedwidth
\newcommand\whline{\noalign{\global\savedwidth\arrayrulewidth
		\global\arrayrulewidth 1.25pt}%
	\hline
	\noalign{\global\arrayrulewidth\savedwidth}}

\ifCLASSINFOpdf
\else
\fi

\definecolor{darkblue}{rgb}{0.0,0.0,1.0}
\hypersetup{colorlinks,breaklinks,
	linkcolor=darkblue,urlcolor=darkblue,
	anchorcolor=darkblue,citecolor=darkblue}

\begin{document}
\setul{}{1.5pt}

\title{Congested Crowd Instance Localization with Dilated Convolutional Swin Transformer}

\author{
        Junyu~Gao,~\IEEEmembership{Member,~IEEE,}
        ~Maoguo~Gong,~\IEEEmembership{~Senior Member,~IEEE,}
        and Xuelong Li,~\IEEEmembership{~Fellow,~IEEE}
	\thanks{

	J. Gao is with the Academy of Advanced Interdisciplinary Research,
	Xidian University, Xi’an 710071, Shaanxi, China, and the School of Artificial Intelligence, Optics and Electronics (iOPEN), Northwestern Polytechnical University, Xi'an {\rm 710072}, P. R. China. E-mail: gjy3035@gmail.com.
	
	M. Gong is with the Key Laboratory of Intelligent Perception and Image Understanding of Ministry of Education, International Research Center for Intelligent Perception and Computation, Xidian University, Xi’an 710071, Shaanxi, China. E-mail: gong@ieee.org.
	
	X. Li is with the School of Artificial Intelligence, Optics and Electronics (iOPEN), Northwestern Polytechnical University, Xi'an {\rm 710072}, P. R. China. E-mail: li@nwpu.edu.cn.
    }
}
\markboth{{IEEE} Transactions on XXX}%
{Shell \MakeLowercase{\textit{et al.}}: Bare Demo of IEEEtran.cls for Journals}
\maketitle

\begin{abstract}

Crowd localization is a new computer vision task, evolved from crowd counting. Different from the latter, it provides more precise location information for each instance, not just counting numbers for the whole crowd scene, which brings greater challenges, especially in extremely congested crowd scenes. In this paper, we focus on how to achieve precise instance localization in high-density crowd scenes, and to alleviate the problem that the feature extraction ability of the traditional model is reduced due to the target occlusion, the image blur, etc. To this end, we propose a Dilated Convolutional Swin Transformer (DCST) for congested crowd scenes. Specifically, a window-based vision transformer is introduced into the crowd localization task, which effectively improves the capacity of representation learning. Then, the well-designed dilated convolutional module is inserted into some different stages of the transformer to enhance the large-range contextual information. Extensive experiments evidence the effectiveness of the proposed methods and achieve the state-of-the-art performance on five popular datasets. Especially, the proposed model achieves F1-measure of 77.5\% and MAE of 84.2 in terms of localization and counting performance, respectively. 

\end{abstract}


\section{Introduction}
\label{intro}

Recently, crowd localization is a hot topic in the field of crowd analysis due to more accurate prediction results than other tasks, such as crowd counting \cite{DBLP:conf/iccv/LiuQLLOL19,gao2020feature,wan2021fine,cheng2021decoupled,Wan_2021_CVPR} and flow estimation \cite{rao2015crowd,DBLP:conf/aaai/LinFLLJ19}. It takes individuals in the crowd scene as the basic unit instead of the scenes. Instance-level localization yields each person's position, which may assist other crowd analysis semantic tasks, trajectory prediction \cite{DBLP:conf/cvpr/AlahiGRRLS16}, anomaly detection \cite{yuan2014online,lin2021learning}, video summarization \cite{liVideoDistillation2021,DBLP:journals/corr/abs-2105-04066}, group detection \cite{li2017multiview,li2020quantifying,DBLP:journals/pami/WangCNL20}, \emph{etc.} Accurate head localization is important for tracking it, predicting its trajectory, recognizing its action and other high-level tasks. Thus, crowd localization is also a fundamental task in crowd analysis. 

\begin{figure}[t]
	\centering
	\includegraphics[width=0.5\textwidth]{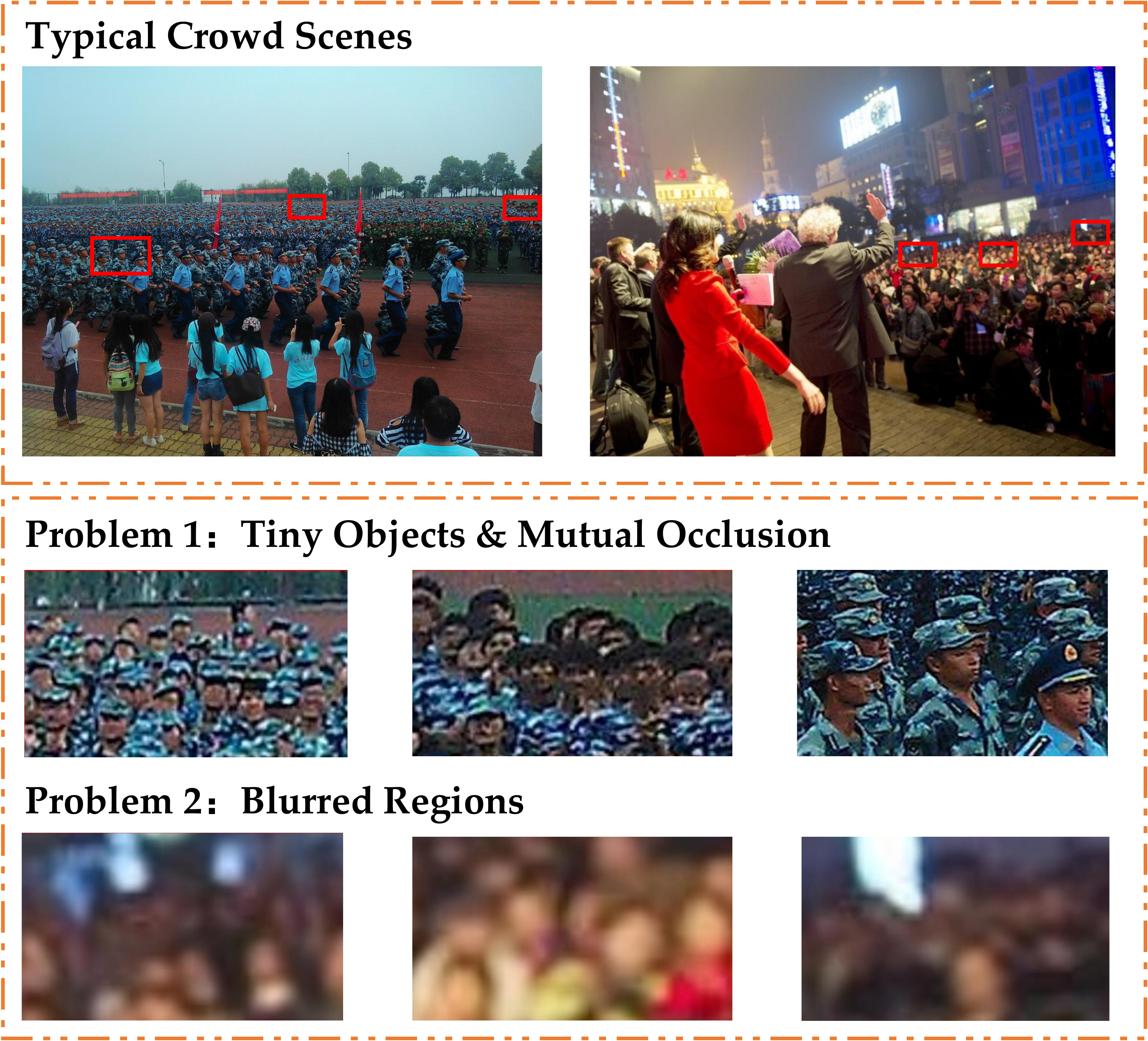}
	\caption{Two problems in some typical crowd scenes: Tiny Objects \& Mutual Occlusion and Blurred Regions, which cause that the current crowd localization models perform poorly in dense crowd scenes. }
	\label{fig:intro}
\end{figure}

\subsection{Motivation}

Currently, there are many researchers focus on crowd localization task. Benefited from the development of the object detection and localization \cite{ren2015faster,DBLP:conf/cvpr/RedmonDGF16}, some methods are proposed for person/head/face detection in sparse scenes, namely low-density crowds. Representative algorithms are DecideNet \cite{liu2018decidenet}, TinyFaces \cite{hu2017finding}, and so on. To handle the dense crowds, some methods \cite{sam2020locate,wang2021self} exploit point supervision to train a locator. Unfortunately, they perform not well for large-scale heads because of a lack of scale information. To reduce the scale-invariant problem, Gao \emph{et al.} \cite{DBLP:journals/corr/abs-2012-04164} propose a segmentation-based localization framework, which treats each head as a non-overlapped instance area and directly outputs its independent semantic head region.

However, in extremely congested scenes, the traditional models can not work well. The main reasons are: 1) small-scale objects and mutual occlusion lack detailed appearance; 2) blurred crowd regions lead to missing the structured patterns of faces. This situation generally occurs at the far end of the shooting angle of view. Fig. \ref{fig:intro} demonstrates the aforementioned issues using two typical crowd scenes. For alleviating these two problems, this paper proposes a high-capacity Dilated Convolutional Swin Transformer (DCST) for instance localization in the extremely congested crowd scenes. For the first problem, we exploit a popular vision transformer, Swin Transformer (ST) \cite{liu2021swin},  to encode richer features than traditional CNN. Then by re-organizing the feature to spatial level, the model can output independent instance map (IIM) \cite{DBLP:journals/corr/abs-2012-04164}. Finally, the FPN decoder \cite{DBLP:conf/cvpr/LinDGHHB17} is added to ST to produce a segmentation map with the same size as input.  

For the second problem, we present a method of modeling the context is used to assist in estimating instance locations in the blurred regions. Although Swin Transformer adopts shifted window to enlarge respective fields in different layers, we find that ST+FPN also performs poorly. The effect of this operation on the encoding of context information is limited. Thus, we attempt to add traditional dilated convolutional layers to the different stages in Swin Transformer, named as ``Dilated Convolutional Swin Transformer'', DCST for short. Specifically, the dilatation module is designed, which consists of two convolutional layers with the dilated rate 2 and 3, respectively. After a stage in the transformer, the features are re-ordered like CNN's feature map. Then by a proposed dilatation module, the large-range contextual information is successfully encoded. Finally, the feature map is recalled with the original size and fed into the next stage in the transformer. Compared with the original Swin Transformer, DCST has a larger respective field and can learn  contextual information more effectively.

\subsection{Contributions}

In summary, the contributions of this paper are three-fold:

\begin{enumerate}
	\item[1)] Propose an effective framework for crowd localization, consisting of a vision transformer as the encoder and an FPN as the decoder.  
	\item[2)] Design a flexible dilatation module and insert it in the transformer encoder, which prompts the contextual encoding capability.  
	\item[3)] The proposed DCST achieves the state-of-the-art performance on the six benchmark or datasets, NWPU-Crowd, JHU++, UCF-QNRF, ShanghaiTech A/B, and FDST.
 	
\end{enumerate}

\subsection{Organization}

The rest of this paper is organized as follows. Section \ref{related} briefly lists and reviews the related literature and works about crowd localization and transformer. Then, Section \ref{method} describes the proposed Dilated Convolutional Swin Transformer (DCST) framework for independent instance segmentation and network architecture. Further, Section \ref{exp} conducts the extensive experiments, and Section \ref{discuss} further analysis the key settings of the method. Finally, this work is summarized in Section \ref{conclusion}.

\section{Related Works}

\label{related}
In this section, the related works about crowd localization and vision transformer are briefly reviewed.

\subsection{Crowd Localization}

\textbf{Detection-based models\,\,\,\,} In the early days, few methods directly focus on individual localization in crowd scenes. Most algorithms try to detect pedestrians, heads, faces, \emph{etc.} in natural images. Specifically, Liu \emph{et al.} \cite{liu2005detecting} propose a segmentation-based method to detect individuals in Surveillance Applications. Andriluka \emph{et al.} \cite{andriluka2009pictorial} present a non-rigid object detection framework, which is based on pictorial structures model \cite{felzenszwalb2005pictorial} and strong part detectors \cite{andriluka2008people} for people detection. Considering the occlusion problem in crowded scenes, some methods focus on detecting the head to locate each individual. Rodriguez \emph{et al.} \cite{rodriguez2011density} propose a density-aware head detection algorithm, which effectively leverages information on the scene's global structure and resolved all detections. Van \emph{et al.} \cite{van2011head} design a template matching method using point cloud data for head detection. Stewart and Andriluka \cite{stewart2016end} present an end-to-end detector based on OverFeat \cite{sermanet2013overfeat} to locating the head position. In addition to the person and head localization algorithms, some detection methods aim to detect tiny faces in dense crowds. Hu and Ramanan \cite{hu2017finding} propose a tiny face detection method, which explores the impacts of image resolution, face scale, \emph{etc.} Li \emph{et al.} \cite{2019PyramidBox} design a context-based module in face detector and propose a data augmentation strategy (Data-anchor-sampling) to prompt the performance. Deng \emph{et al.} \cite{deng2019retinaface}  annotate five landmarks for each face to improve the performance of hard sample localization.

\textbf{Point-based models\,\,\,\,} The aforementioned detection-based methods are not suitable for dense crowd scenes, especially when there are more than 1,000 people. Before 2020, the common congested crowd datasets do not provide box-level annotations. Thus, some point-based methods are very popular in this field. Idrees \emph{et al.} \cite{idrees2018composition} attempt to find the peak point in the predicted density maps. By locating the maximum value in a local region, the head position is obtained. Liu \emph{et al.} \cite{liu2019recurrent} propose a new label type, Cruciform, which easier to be located maximum value than traditional density maps. Gao \emph{et al.} \cite{gao2019domain} design an iterative scheme to find maximum values in reverse for density maps. Wan and Chan \cite{wan2020modeling} present a new method to construct the noises of point annotations, which enhances the robustness of the crowd model.  Wang \emph{et al.} \cite{wang2021self} present a self-training mechanism based on a key-point detector to predict the head center. Wang \emph{et al.} \cite{wang2021dense} construct a baseline for crowd localization based on a point-based detection method \cite{zhou2019objects}. Sam \emph{et al.} \cite{sam2020locate} propose a detector tailor-made for dense crowds only relying on pseudo box labels generated by point information. Liang \emph{et al.} \cite{liang2021focal} design a Focal Inverse Distance Transform map to depict labels, and propose an I-SSIM loss to detect local Maxima. Wan \emph{et al.} \cite{Wan_2021_CVPR} propose a generalized loss function to learn robust density maps for counting and localization simultaneously.

\textbf{Segmentation-based models\,\,\,\,} With the release of high-resolution datasets, NWPU-Crowd \cite{gao2020nwpu}, segmentation-based methods attract many researchers' attention. Abousamra \emph{et al.} \cite{abousamra2020localization} propose a topological constraint to model the spatial arrangement, which uses a persistence loss based on the persistent homology. Gao \emph{et al.} \cite{gao2020learning} propose an adaptive threshold module to elaborately segment tiny heads in dense crowd regions. Considering that segmentation maps provide a more refined and reasonable label, this paper will be based on it to deploy our works. 

\begin{figure*}[t]
	\centering
	\includegraphics[width=1\textwidth]{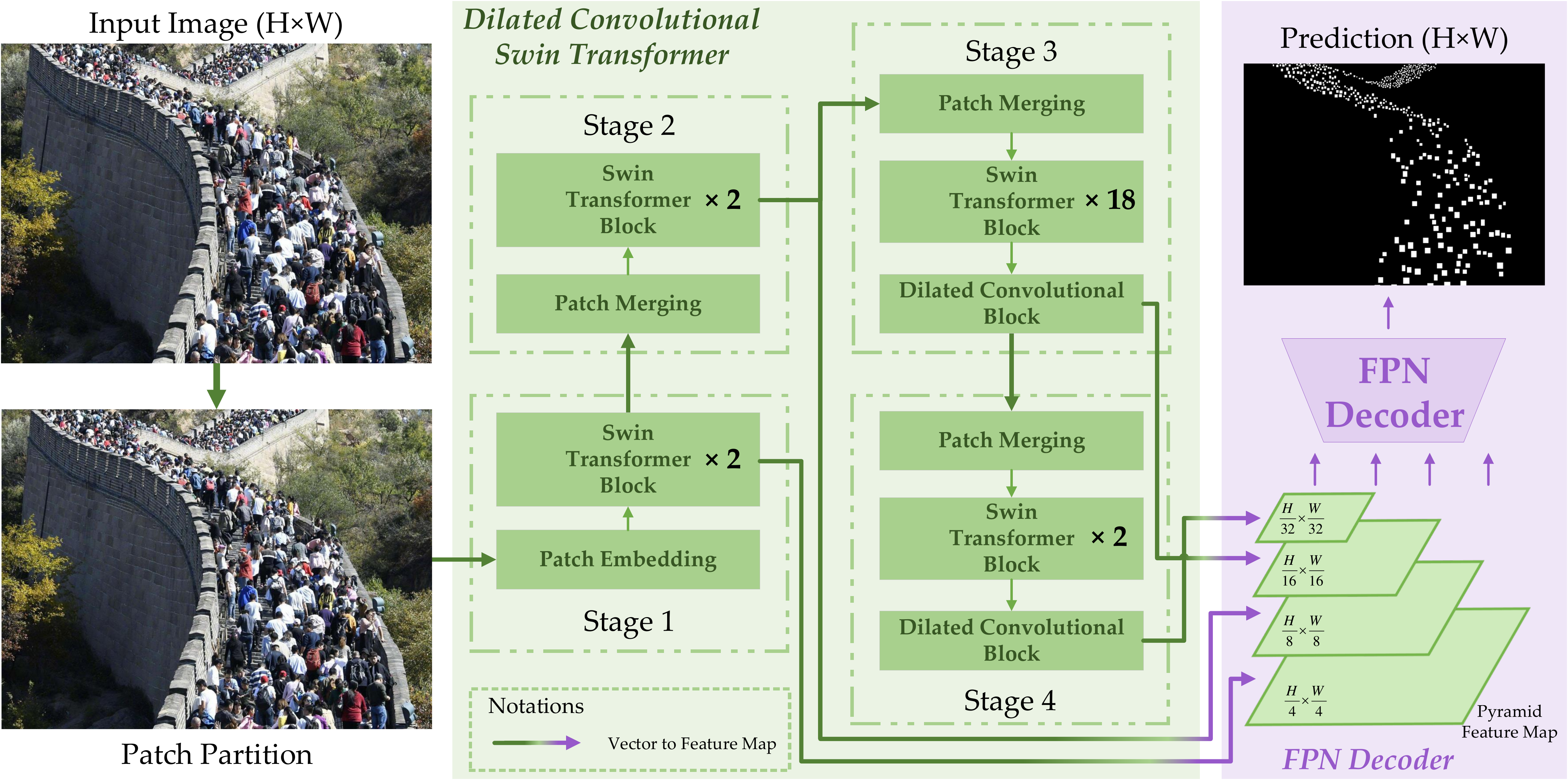}
	\caption{The flowchart of the proposed Dilated Convolutional Swin Transformer and FPN decoder (DCST+FPN). Notably, In Stage 3 and 4, the Dilated Convolutional Block (DCB) is applied to the top of the Swin Transformer Block (STB). By a Feature Pyramid Networks (FPN) decoder, the features from different stages are fused, and the final independent instance segmentation map with the input size is produced. }
	\label{fig:framework}
\end{figure*}

\subsection{Vision Transformer}

Transformer is proposed by Vaswani \emph{et al.} \cite{vaswani2017attention}. It is widely used in many NLP tasks due to its powerful feature extraction capacity. In 2020,  
Dosovitskiy \emph{et al.} \cite{dosovitskiy2020image} propose a vision transformer (ViT) for image recognition, which shows the high-performance representation learning ability in computer vision tasks. After this, many vision transformer variants are presented. Yuan \emph{et al.} \cite{yuan2021tokens} present a Tokens-to-Token ViT (T2T ViT), which can encode the local structures for each token. Wang \emph{at al.} \cite{wang2021pyramid} propose a backbone transformer for dense prediction, named as ``Pyramid Vision Transformer (PVT)'',which designs a shrinking pyramid scheme to reduce the traditional transformer's sequence length. Han \emph{et al.} \cite{han2021transformer} propose a Transformer-iN-Transformer (TNT) architecture, of which the inner extracts local features, and the outer processes patch embeddings. To achieve a trade-off between speed and accuracy, Liu \emph{et al.} \cite{liu2021swin} introduce a shifted window strategy into transformer to encode representations. In the field of crowd analysis, Liang \emph{et al.} \cite{liang2021transcrowd} propose a transformer for weakly-supervised counting, which exploits a transformer to directly regress the number of counting. Sun \emph{et al.} \cite{Sun2021boosting} design a token-attention module to encode features by channel-wise attention, and a regression-token module to produce the count of people in crowd scenes.

\section{Approach}
\label{method}

This section firstly reviews the basic Vision Transformer (ViT) and its variants Shift Window ViT (ST).  Then we describe the proposed Dilated Convolutional Shift Window ViT (DCST). Finally, the network architecture, loss function, and implementation details are reported. 

\subsection{Vision Transformer (ViT)}

At present, vision transformer shows its powerful capacity of representation learning. In 2017, transformer is proposed by Vaswani \emph{et al.} \cite{vaswani2017attention} and become a standard operation in the field of Natural Language Processing (NLP). Notably, BERT \cite{devlin2018bert} and GPT \cite{brown2020language} obtain remarkable progress in the related tasks. Carion \emph{et al.} \cite{carion2020end} exploit transformer to detect objects, which is added to the top of traditional CNNs. Dosovitskiy \emph{et al.} \cite{dosovitskiy2020image} propose a fully transformer architecture for image classification, named as ``Vision Transformer (ViT)''. Here, we briefly review ViT.

\subsubsection{Patch embeddings} 
Given an image $i \in {\mathcal{R}^{H \times W \times C}}$ with the size of height $H$, width $W$ and channel $C$ , it is reshaped into a sequence, which consists of $N$ ($N = {{HW} \mathord{\left/{\vphantom {{HW} {{P^2}}}} \right.\kern-\nulldelimiterspace} {{P^2}}}$) patches ${i_p}$ with the size of $P \times P \times C$. Then each patch is flattened and mapped to $D$ dimensions latent vector using a trainable linear projection  in transformer, of which output is named as ``patch embeddings''. In addition, position embeddings ${\bm{E}_{pos}}$ are added to the patch embeddings to represent positional information, which are learnable 1D embeddings. Finally, The sequence of embedding vectors are fed into the Transformer Encoder. 

Specifically, the operation of patch embeddings is formulated as follows:

\begin{equation}
\begin{array}{l}
\begin{aligned}
{z_0} = [{i_{class}};\;i_p^1\bm{E};\;i_p^2\bm{E};\;...;\;i_p^N\bm{E}] + {\bm{E}_{pos}},
\end{aligned}\label{E}
\end{array}
\end{equation}
where ${i_{class}}$ is the embedded patches $z_0^0$, and $\bm{E}$ denotes the process of the learnable embeddings ($\bm{E} \in {R^{({P^2} \times C) \times D}}$, $\bm{E} \in {R^{(N + 1) \times D}}$). 

\subsubsection{Transformer Encoder} 
Transformer Encoder includes Multi-headed Self-Attention (MSA) and Multi-Layer (MLP) modules. Given a $L$ layers of Transformer Encoder, MSA and MLP are formulated as: 

\begin{equation}
\begin{array}{l}
\begin{aligned}
z_l^{\prime} = MSA(LN({z_{l - 1}})) + {z_{l - 1}},\quad l = 1,...,L,
\end{aligned}\label{MSA}
\end{array}
\end{equation}

\begin{equation}
\begin{array}{l}
\begin{aligned}
{z_l} = MLP(LN(z_l^{\prime})) + z_l^{\prime},\quad l = 1,...,L,
\end{aligned}\label{MLP}
\end{array}
\end{equation}
where LN denotes Layer Normalization \cite{DBLP:journals/corr/BaKH16} for stable training. In MLP, two layers with GELU non-linearity activation function \cite{Hendrycks} are applied. Notably, LN is employed for each sample $z$:

\begin{equation}
\begin{array}{l}
\begin{aligned}
LN(z) = \frac{{z - \mu }}{\delta }  \circ \gamma  + \beta,
\end{aligned}\label{LN}
\end{array}
\end{equation}
where $\mu$ and $\delta$ denote the mean and standard deviation of features respectively, $\gamma$ and $\beta$ are the learnable parameters of affine transformation, and $\circ$ is element-wise dot operation.

\subsection{Swin Transformer}

Compared with ViT, Swin Transformer is a hierarchical architecture for handling dense prediction problems and reducing the computational complexity. Specifically, it computes self-attention in non-overlapping windows with small-scale sizes. Further, to encode contextual information, the window partitions in consecutive layers are different. Consequently, the large-range information is transformed in the entire network by local self-attention modules.

Swin Transformer contains four stages to produce different number tokens. Given an image with the size of $H \times W$, token is a raw pixels concatenation vector of an RGB image patch with the size of $4 \times 4$. A linear embedding is employed on this token to map it into a vector with the dimension $C$. Stage 1, 2, 3, and 4 produce $\frac{H}{4} \times \frac{W}{4}$, $\frac{H}{8} \times \frac{W}{8}$, $\frac{H}{16} \times \frac{W}{16}$, and $\frac{H}{32} \times \frac{W}{32}$ tokens, respectively. Each stage consists of a Patch Embedding and some Swin Transformer Blocks. Different from MSA in ViT, a Swin Transformer Blocks uses the  shifted-window MSA to compute locally self-attention.

\subsection{Dilated Convolutional Swin Transformer}

Although Swin Transformer design a shifted-widow scheme of the sequential layers in a hierarchical architecture, large-range spatial contextual information is still encoded not well. For alleviating this problem, we propose a Dilated Convolutional Swin Transformer (``DCST'' for short) to enlarge the respective field for spatial images. In this way, the large-range contextual information can be encoded well on different scales. To be specific, the Dilated Convolutional Block is designed and inserted into between different stages of Swin Transformer.

\textbf{Dilated Convolution\,\,\,\,} Dilated Convolution is proposed by Yu and Koltun \cite{yu2015multi} in 2015. Compared with the traditional convolution operation, the dilated convolution supports the expansion of the receptive field. Notably, a traditional $3 \times 3$-kernel convolution has a respective field of $3 \times 3$. If it is a 2-dilated convolution with the same kernel size, the respective field is $7 \times 7$. Thus, the dilated convolution can expand the respective field without loss of feature resolution.

\textbf{Dilated Convolutional Block (DCB)\,\,\,\,}  Considering that the data flow in Swin Transformer is vectors instead of feature maps in traditional CNNs, DCB firstly reshapes a group of vector features into a spatial feature map. For example, the number of $\frac{H}{4} \times \frac{W}{4}$ $C$-dimension tokens is reshaped as a feature map with the size of $\frac{H}{4} \times \frac{W}{4} \times C$. After this, two dilated convolutional with Batch Normalization \cite{ioffe2015batch} and ReLU are applied to extract large-range spatial features. Finally, the feature map is re-transformed the original number and size (namely the number of $\frac{H}{4} \times \frac{W}{4}$ $C$-dimension tokens) and fed into the next stage in Swin Transformer.

\subsection{Network Configurations}

For the dense prediction task, a classic architecture is an Encoder-Decoder network to output results with the same size of the inputs. In this paper, the encoder is the proposed DCST and the decoder is based on FPN \cite{lin2017feature}.

\textbf{Encoder: DCST\,\,\,\,} In DCST, the Swin Transformer is Swin-B, of which four stages has 2, 2, 18, and 2 Swin Transformer Blocks. After Stage 3 and 4, the Dilated Convolutional Block (DCB) is added. Following by \cite{wang2018understanding}, the dilatation rate of the two dilated convolutional layers in DCB is $2$ and $3$. 

\textbf{Decoder: FPN\,\,\,\,} Similar to IIM \cite{DBLP:journals/corr/abs-2012-04164}, this paper also utilizes FPN \cite{lin2017feature} to fuse different-scale features. To be specific, for DCST's four stages, the four-head FPN is designed. Finally, for a high-resolution output to obtain an independent instance map, a convolutional layer and two de-convolutional layers is applied to yields the $1$-channel feature map with the original input size. A sigmoid activation is employed to normalize the results in $(-1,1)$, which is named as ``score map''.   

\subsection{Loss Function}

For training the crowd localization model, we adopt the standard Mean Squared Error (MSE) loss function.

\subsection{Implementation Details}

\textbf{Training Setting \,\,\,\,} For augmenting the data, random horizontally flipping, random scaling (target scales: $0.8 \times \sim1.2 \times$ original scales) and random cropping (target size: 512px $\times$ 1024px) are employed. The experiments are conducted on two NVIDIA TITAN RTXs ($\sim$48GB GPU Memory) The batch size is $8$. The base learning rate of transformer and FPN is set as $0.6 \times {10^{ - 5}}$, and the dilated conv layers' base learning rate is set as $0.6 \times {10^{ - 6}}$. A linear warmup is implemented in first 1,500 iterations. AdamW \cite{loshchilov2017decoupled} algorithm is utilized to optimize the proposed network. 

\textbf{Threshold Selection \,\,\,\,} For the score map yielded by the network, we need to select a proper threshold to transform it into a binary map. The purpose is that each independent instance can be detected by connected component segmentation. Notably, the best threshold is selected on the validation set and is directly applied on the test set.

\section{Experimental Results}

\label{exp}

The experiments are conducted on six main-stream crowd datasets, including NWPU-Crowd, JHU-CROWD++, FDST, UCF-QNRF, and ShanghaiTech Part A/B Dataset. At the same time, the further analyses are discussed in Section \ref{discuss} on the NWPU-Crowd \emph{validation set}.

\subsection{Evaluation Criteria}

\label{metrics}

\textbf{Crowd Localization\,\,\,\,} In this section, we follow NWPU-Crowd \cite{gao2020nwpu} to evaluate instance-level Precision, Recall, and F1-measure (Pre., Rec., and F1-m for short in the following tables), which are calculated under the adaptive scale for each head. To be specific, the definitions of the above are:

\begin{equation}
\begin{array}{l}
\begin{aligned}
precision = \frac{{TP}}{{TP + FP}},
\end{aligned}
\end{array}
\end{equation}

\begin{equation}
\begin{array}{l}
\begin{aligned}
recall = \frac{{TP}}{{TP + FN}},
\end{aligned}
\end{array}
\end{equation}

\begin{equation}
\begin{array}{l}
\begin{aligned}
{F_1} = \frac{{\left( {{\beta ^2} + 1} \right) \cdot precision \cdot recall}}{{{\beta ^2} \cdot precision + recall}},\quad \beta  = 1,
\end{aligned}
\end{array}
\end{equation}
where TP, FP, FN denote the number of True Positive, False Positive, and False Negative, respectively. The TP, FP, and FN is calculated under large $\sigma$, namely the radius of circumcircle for a head box label.

\textbf{Crowd Counting\,\,\,\,} In addition to the metrics for localization, we also evaluate the counting performance using Mean Absolute Error (MAE), Mean Squared Error (MSE), and mean Normalized Absolute Error (NAE), which are defined as:

\begin{equation}
\begin{array}{l}
\begin{aligned}
MAE = \frac{1}{N}\sum\limits_{i = 1}^N {\left| {{y_i} - {{\hat y}_i}} \right|}
\end{aligned}
\end{array}
\end{equation}

\begin{equation}
\begin{array}{l}
\begin{aligned}
MSE = \sqrt {\frac{1}{N}\sum\limits_{i = 1}^N {{{\left| {{y_i} - {{\hat y}_i}} \right|}^2}} },
\end{aligned}
\end{array}
\end{equation}

\begin{equation}
\begin{array}{l}
\begin{aligned}
NAE = \frac{1}{N}\sum\limits_{i = 1}^N {\frac{\left| {{y_i} - {{\hat y}_i}} \right|}{{y_i}}},
\end{aligned}
\end{array}
\end{equation}
where N is the number of samples in test or validation set, $y_i$ is GT number and ${{{\hat y}_i}}$ is predicted number for $i$-th sample.

\subsection{Datasets}
\label{dataset}
\textbf{NWPU-Crowd} \cite{gao2020nwpu} is a large-scale high-quality and high-resolution crowd counting and localization dataset, consisting of 5,109 scenes and 2,133,238 instances. It provides two types of annotation, point-level and box-level for each head. 

\textbf{JHU-CROWD++} \cite{sindagi2020jhu-crowd++} is an extension version of JHU-CROWD \cite{sindagi2019pushing}, which consists of 4,372 images, $\sim$1.5 million instances. It contains high-diversity crowd scenes, such as rain, snow, haze, \emph{etc.} For each head, the point level, approximate size, blur level, occlusion level are labeled. 

\textbf{FDST} \cite{fang2019locality} is a video crowd counting dataset, which consists of 13 different scenes, 100 image sequences, 150,000 frames. It also annotates point-level and box-level labels simultaneously. 

\textbf{UCF-QNRF} is proposed by Idrees \emph{et al.} \cite{idrees2018composition}, which is a congested crowd counting dataset, consisting of 1,535 dense scenes, with a total of 1,251,642 instances. The average of counting number is $\sim815$.

\textbf{Shanghai Tech} is proposed by Zhang \emph{et al.} \cite{zhang2016single} , which has two subsets, Part A and B. The former labels 482 images, 241,677 instances; the latter has 716 images, including 88,488 labeled heads. 

Note that UCF-QNRF and Shanghai Tech datasets do not provide the head scale. Following Gao \emph{et al.} \cite{DBLP:journals/corr/abs-2012-04164}, we adopt the predicted scale information to evaluate the performance for localization. The generation code for scale information is available at \url{https://github.com/taohan10200/IIM}.

\subsection{Ablation Study on the NWPU-Crowd validation set}

To further understand each component in the proposed method, this section compares the step-wise results on the NWPU-Crowd \emph{validation set}. 

\textbf{ST}: A crowd counting model. Swin Transformer for image recognition model. The last classification layer is replaced by a sigmoid layer to directly output the number of people in crowd scenes.

\textbf{DCST}: A crowd counting model. Dilated Convolutional Swin Transformer for image recognition model. Similarly, the last classification layer is replaced by a Sigmoid layer.

\textbf{ST+base decoder}: A crowd counting and localization model. Swin Transformer is used as a backbone to extract features. The decoder is added to the top of Stage 4 in Swin Transformer, which consists of one convolutional layer and two de-convolutional layer output high-resolution score maps.

\textbf{ST+FPN}: Different from ST+base decoder, FPN decoder is employed to fuse the outputs of four stages in Swin Transformer and to output high-quality score maps.

\textbf{DCST+FPN}: Compared with ST+FPN, the backbone is replaced by DCST.

\begin{table}[htbp]	
	\centering
	\caption{Ablation study of different components in the proposed method on the NWPU-Crowd \emph{validation set}. }
	\setlength{\tabcolsep}{1.4mm}{\begin{tabular}{cIc|c|cIc|c|c}
			\whline
			\multirow{2}{*}{Method}  &\multicolumn{3}{cI}{Localization} &\multicolumn{3}{c}{Counting} \\
			\cline{2-7}
			&\textbf{F1-m} & Pre & Rec &MAE &MSE&NAE	\\
			\whline
			ST     &- &- & - & 103.2 & 393.2  & 0.307	\\
			\hline
			DCST     &-   &-   &- & 96.3  & 412.0 & 0.248 	\\
			\whline
			ST+base decoder           &73.7  & 80.1 & 68.2 & 77.5 & 324.0  &0.193	\\
			\hline
			ST+FPN          & 79.8& 88.3 & 72.7 &  54.6&181.4 & 0.159  	\\
			\hline
			DCST+FPN        &\textbf{81.4}  & \textbf{84.1}   & \textbf{79.0} &\textbf{43.9}  &\textbf{109.8}  & \textbf{0.156}\\
			\whline			
	\end{tabular}}
	\label{Table:ablation}
\end{table}

\begin{figure*}[t]
	\centering
	\includegraphics[width=1\textwidth]{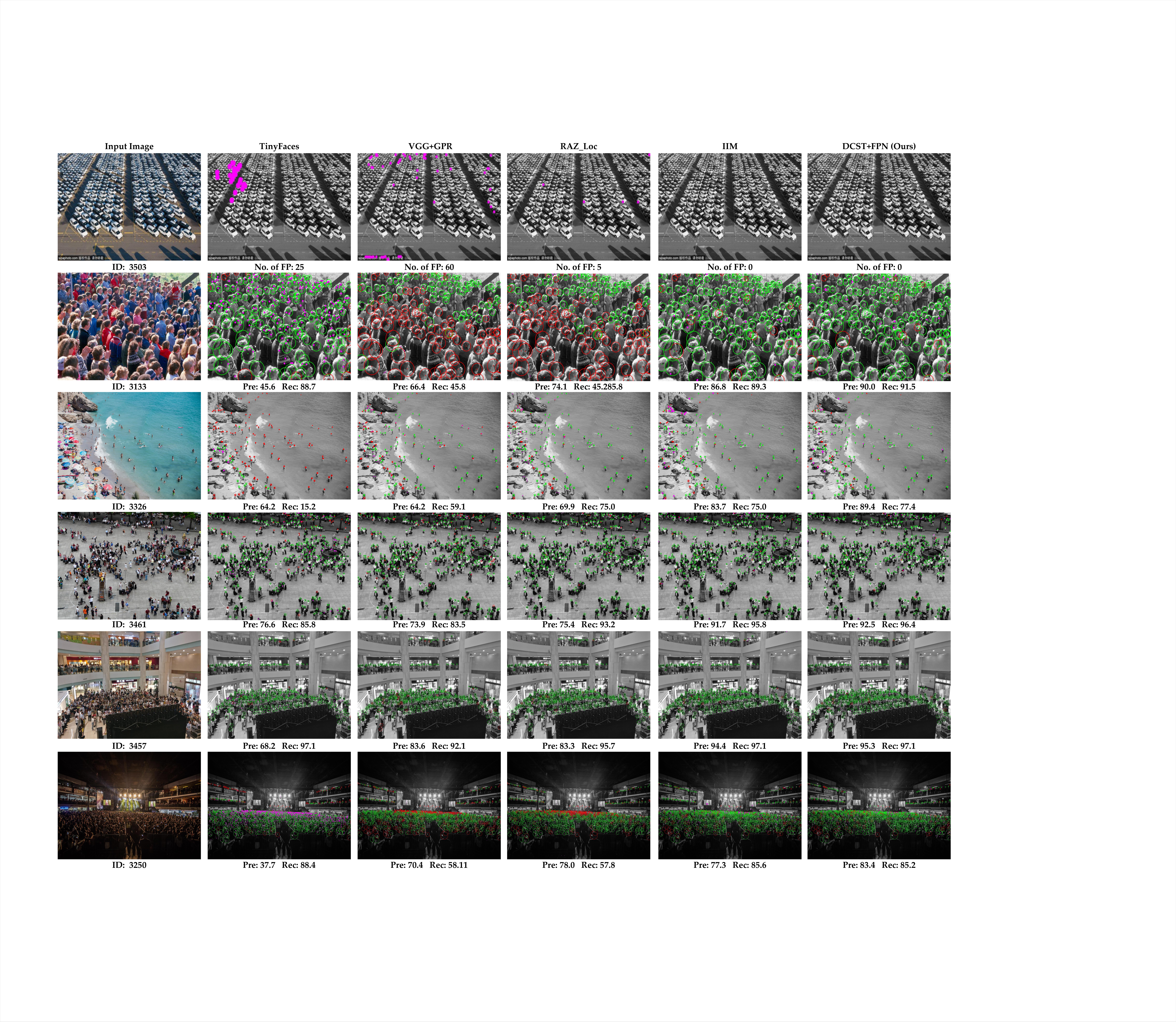}
	\caption{Six groups of visualization predicted results of four popular methods and the proposed DCST+FPN on NWPU-Crowd \emph{validation set}. The showing  contains Negative Samples (ID: 3503), large-scale heads (ID: 3133), tiny heads (ID: 3326 \& 3461), dense crowds (ID: 3457 \& 3250) and low-light scenes(ID: 3250). For a better visual effect, predicted images' contents are shown by the gray scale, where {\color{green}{green}}, {\color{red}{red}} and {\color{magenta}{magenta}} points denote true positive (TP), false negative (FN) and false positive (FP), respectively. Besides, the {\color{green}{green}} and {\color{red}{red}} circles are GroundTruth with the radius of $\sigma_l$.}
	\label{fig:re}
\end{figure*}

Table \ref{Table:ablation} lists the performance of each model for localization and counting on the NWPU-Crowd \emph{validation set}. Considering that ST and DCST directly output the number of people in crowd scenes, the localization metrics are not reported. From the table, the full model (DCST+FPN) achieves better performance (F1-measure of 81.4 and MAE of 43.9) than other baseline methods. From the overall trend, after the introduction of new modules, the performance is improved significantly. Besides, we find that pixel-wise-based methods have better results compared with scene-level-based algorithms (Row 3,4,5 \emph{v.s.} Row 1,2). The former provides spatial annotations in the training process, but the latter only directly regresses a number. Therefore, Row 3,4,5's methods learn more reasonable features than Row 1,2.

\textbf{Effect of the DCB.\,\,} By comparing two groups of experiments (ST \emph{v.s.} DCST and ST+FPN \emph{v.s.} DCST+FPN), we find that introducing Dilated Convolutional Block (DCB) in Swin Transformer can effectively prompt the performance of localization and counting. Specifically, the estimation errors (MAE) increases by 6.7\% and 19.6\% for counting. At the same time, there is an interesting phenomenon that DCB is more effective in pixel-wise training than scene-level training (19.6\% \emph{v.s.} 6.7\%), which evidences the dilated convolution captures the large-range contextual features for dense prediction tasks.

\begin{table*}[htbp]
	\centering
	
	\caption{The leaderboard of NWPU-Crowd Localization (\emph{test set}), excluding some anonymous submissions. The bold and underlined text denotes the first and second place, respectively. Avg.[B] of Recall means the average of recalls for different box level $[10^0,10^1]$, $(10^1,10^2]$,  $(10^2,10^3]$,  $(10^3,10^4]$,  $(10^4,10^5]$, and $>10^5$. The timestamp is Jun. 28, 2021, which is recorded in the official CrowdBenchmark website (the link is \url{https://www.crowdbenchmark.com/historyresultl.html}).}
	\begin{tabular}{cIcIcIc|cIc}
		\whline
		\multirow{2}{*}{Method}	&\multirow{2}{*}{Venue} &\multirow{2}{*}{Backbone}  &\multicolumn{2}{cI}{Overall ($\sigma_l$)}  &Recall ($\sigma_l$)   \\
		\cline{4-6}
		& &  & \textbf{F1-m}/Pre/Rec (\%) $\uparrow$  &MAE/MSE/NAE $\downarrow$ & Avg.[B] (\%) $\uparrow$ \\
		\whline
		Faster R-CNN \cite{ren2015faster}   & ICCV'15& ResNet-101  & 6.7/\textbf{95.8}/3.5 &414.2/1063.7/0.791 &18.2  \\
		\hline
		TinyFaces \cite{hu2017finding} &CVPR'17 &ResNet-101   &56.7/52.9/61.1  &272.4/764.9/0.750  & 59.8  \\
		\hline
		RAZ\_Loc \cite{liu2019recurrent} & CVPR'19 &VGG-16  &59.8/66.6/54.3 &151.5/634.7/0.305 &42.4   \\
		\hline
		VGG+GPR \cite{gao2019c,gao2019domain} & arxiv'19 &VGG-16  & 52.5/55.8/49.6  &127.3/439.9/0.410  & 37.4  \\
		\hline
		AutoScale\_localization \cite{xu2019autoscale}  & arxiv'19 &VGG-16 &62.0/67.4/57.4 &123.9/515.5/0.304 &48.4    \\
		\hline	
		IIM \cite{DBLP:journals/corr/abs-2012-04164} & arxiv'20 &VGG-16  &73.2/77.9/69.2 &96.1/414.4/0.235 &58.7   \\
		\hline
		IIM \cite{DBLP:journals/corr/abs-2012-04164} & arxiv'20 &HRNet  &\underline{76.2}/81.3/\underline{71.7} &87.1/406.2/\textbf{0.152} &\underline{61.3}   \\
		\hline
		TopoCount \cite{abousamra2020localization} & AAAI'21 &VGG-16  &69.2/68.3/70.1 &107.8/438.5/- &\textbf{63.3}   \\
		\hline
		Crowd-SDNet \cite{wang2021self} & T-IP'21 &ResNet-50  &63.7/65.1/62.4 &-/-/- &55.1    \\
		\hline
		FIDTM \cite{liang2021focal} & arxiv'21 &HRNet  &75.5/79.8/\underline{71.7} &86.0/\textbf{312.5}/0.277 &47.5   \\
		\whline
		DCST+FPN (Ours) & - &DCST  &\textbf{77.5}/\underline{82.2}/\textbf{73.4} &\textbf{84.2}/\underline{374.6}/\underline{0.153} &60.9   \\
		\whline
	\end{tabular}
	\label{table:nwpu}
\end{table*}

\textbf{Visualization analysis.\,\,} Fig. \ref{fig:re} illustrates six groups visual localization results on NWPU \emph{validation set}. In Row 1, a Negative Sample, IIM-based methods (IIM and the proposed DCST+FPN) yields a better performance (no FP) than other methods. For large-scale heads (Row 2, ID: 3133), detection-based TinyFaces produces many false positives. Density-regression-based VGG+GPR and segmentation-based RAZ\_Loc miss quite a few instances, especially large heads in the scenes. The main reason is that IIM-based method provides head region information to assist models to learn semantic scales for each head. Fixed kernel ($15 \times 15$ and $3 \times 3$) is not proper for large heads. Similarly, for Row 3 and 6, containing extremely tiny heads, VGG+GPR and RAZ\_Loc miss quite a few instances. From this, the above two methods work not well in extreme scenes. Further, by comparing Column 5 and 6, we find that DSCT+FPN obtain higher Precision than IIM. Take Row 3 and 6 as examples, DSCT+FPN's number of FP is less than that of IIM significantly, which shows the robustness of the proposed DCST backbone.

\subsection{Leaderboard on the NWPU-Crowd Benchmark}

Table \ref{table:nwpu} lists the leaderboard of NWPU-Crowd Localization and Counting Track. All results are evaluated on the \emph{test set}. Due to the lack of algorithm details, some anonymous methods are not listed in the table.
From the leaderboard, the proposed DCST+FPN achieves the best F1-measure of 77.5\% for localization and the best MAE of counting, where F1-measure and MAE are the primary keys for ranking. In other words, DCST+FPN is number one on the leaderboard. Besides, DCST+FPN also achieves one first place (Recall of 73.4\%) and three second places (Precision of 82.2\%, MSE of 374.6, and NAE of 0.153). In terms of Avg.[B] of Recall (the last column in the table), DCST+FPN obtain 60.9\%, which is the third place in all algorithms, less than TopoCount and IIM. In general, our method is superior to other state-of-the-art methods.

\subsection{Comparison with the SOTAs on Other Datasets}

This section reports the performance of four state-of-the-art methods on other mainstream crowd datasets: TinyFaces \cite{hu2017finding}, RAZ\_Loc \cite{liu2019recurrent}, LSC-CNN \cite{sam2020locate}, and IIM \cite{gao2020learning}. To be specific, TinyFaces is implemented by \footnote{https://github.com/varunagrawal/tiny-faces-pytorch} with the default parameters, RAZ\_Loc  is trained with the code \footnote{https://github.com/gjy3035/NWPU-Crowd-Sample-Code-for-Localization} provided by Wang \emph{et al.} \cite{gao2020nwpu}, and LSC-CNN's results is calculated by inference of official code and models \footnote{https://github.com/val-iisc/lsc-cnn}. IIM's performance comes from the corresponding technical report \cite{gao2020learning}.

Table \ref{table:sota} lists the concrete localization results. From it, we find that DCST+FPN shows strong performance, outperforming other methods on dense crowd datasets in terms of F1-measure metric, namely ShanghaiTech Part A, UCF-QNRF, and JHU-CROWD++. For sparse scenes (ShanghaiTech Part B and FDST), the performance of DCST+FPN is close to that of IIM. This phenomenon shows that the proposed DCST+FPN is more suitable for congested crowd scenes due to its powerful capacity for large-range representation learning. In general, DCST+FPN achieves seven first places and six second places in terms of F1-measure, outperforming other state-of-the-art methods.

\begin{table*}[htbp]
	\centering
	
	\caption{The comparison of four popular methods and DCST+FPN on the four datasets, JHU-CROWD++, FDST, UCF-QNRF, and ShanghaiTech Part A/B. All results are computed under $\sigma_l$, which is defined  in Section \ref{metrics}. The bold and underlined text denotes the first and second place, respectively. }
	\setlength{\tabcolsep}{0.25cm}{
	\begin{tabular}{cIcIcIc|c|cIc|c|cIc|c|c}
		\whline
		\multirow{2}{*}{Method} & \multirow{2}{*}{Venue} & \multirow{2}{*}{Backbone} &\multicolumn{3}{cI}{ShanghaiTech Part A} &\multicolumn{3}{cI}{ShanghaiTech Part B} &\multicolumn{3}{c}{UCF-QNRF} \\
		\cline{4-12}
		&& &\textbf{F1-m} & Pre. &Rec. &\textbf{F1-m} & Pre. &Rec. & \textbf{F1-m} & Pre. &Rec. \\
		\hline
		TinyFaces \cite{hu2017finding} &CVPR'17 &ResNet-101 &57.3 &43.1 &\textbf{85.5} &71.1 &64.7 &79.0 &49.4 &36.3 &\textbf{77.3}   \\
		\hline
		RAZ\_Loc\cite{liu2019recurrent}&CVPR'19 &VGG-16 &69.2 &61.3 &79.5 &68.0  &60.0 &78.3 &53.3 &59.4  &48.3  \\
		\hline	
		LSC-CNN \cite{sam2020locate} &T-PAMI'20 &VGG-16 &68.0&69.6 &66.5 &71.2 &71.7 &70.6 &58.2 &58.6 &57.7 \\
		\hline
		IIM & arxiv'20  &VGG-16 &72.5  &72.6  &\underline{72.5}   &80.2   & 84.9  &76.0   &68.8   &78.2   &61.5      \\
		\hline
		IIM & arxiv'20 &HRNet &\underline{73.9} &\textbf{79.8} &68.7 &\textbf{86.2} &\textbf{90.7} &\underline{82.1} &\underline{72.0} &\textbf{79.3} &65.9  \\
		\whline
		DCST+FPN & - &DCST &\textbf{74.5} &\underline{77.2} &72.1 &\underline{86.0} &\underline{88.8} &\textbf{83.3} &\textbf{72.4} &\underline{77.1} &\underline{68.2}  \\
		\whline
		
	\end{tabular}
\vspace{0.15cm}
}

\setlength{\tabcolsep}{0.25cm}{
\begin{tabular}{cIcIcIc|c|cIc|c|c}
	\whline
	\multirow{2}{*}{Method} & \multirow{2}{*}{Venue} & \multirow{2}{*}{Backbone} &\multicolumn{3}{cI}{JHU-CROWD++} &\multicolumn{3}{c}{FDST}\\
	\cline{4-9}
	&&& \textbf{F1-m} & Pre. &Rec. &\textbf{F1-m} & Pre. &Rec.\\
	\hline
	TinyFaces \cite{hu2017finding}&CVPR'17 &ResNet-101 &- &- &- &85.8 &86.1 &85.4  \\
	\hline
	RAZ\_Loc\cite{liu2019recurrent}&CVPR'19 &VGG-16  &- &-  &- &83.7 &74.4&\textbf{95.8}  \\
	\hline
	IIM & arxiv'20  &VGG-16  &-   &-   &-   &93.1  &92.7  &93.5   \\
	\hline
	IIM & arxiv'20 &HRNet &\underline{62.5} &\underline{74.0} &\underline{54.2} &\textbf{95.5} &\underline{95.3}&\textbf{95.8}  \\
	\whline
	DCST+FPN & - &DCST  &\textbf{64.4} &\textbf{74.9} &\textbf{56.5} & \underline{94.8} &\textbf{95.4}& 94.2 \\
	\whline
\end{tabular}
}
	\label{table:sota}
\end{table*}

\section{Discussions}

\label{discuss}

\begin{figure}[t]
	\centering
	\includegraphics[width=0.5\textwidth]{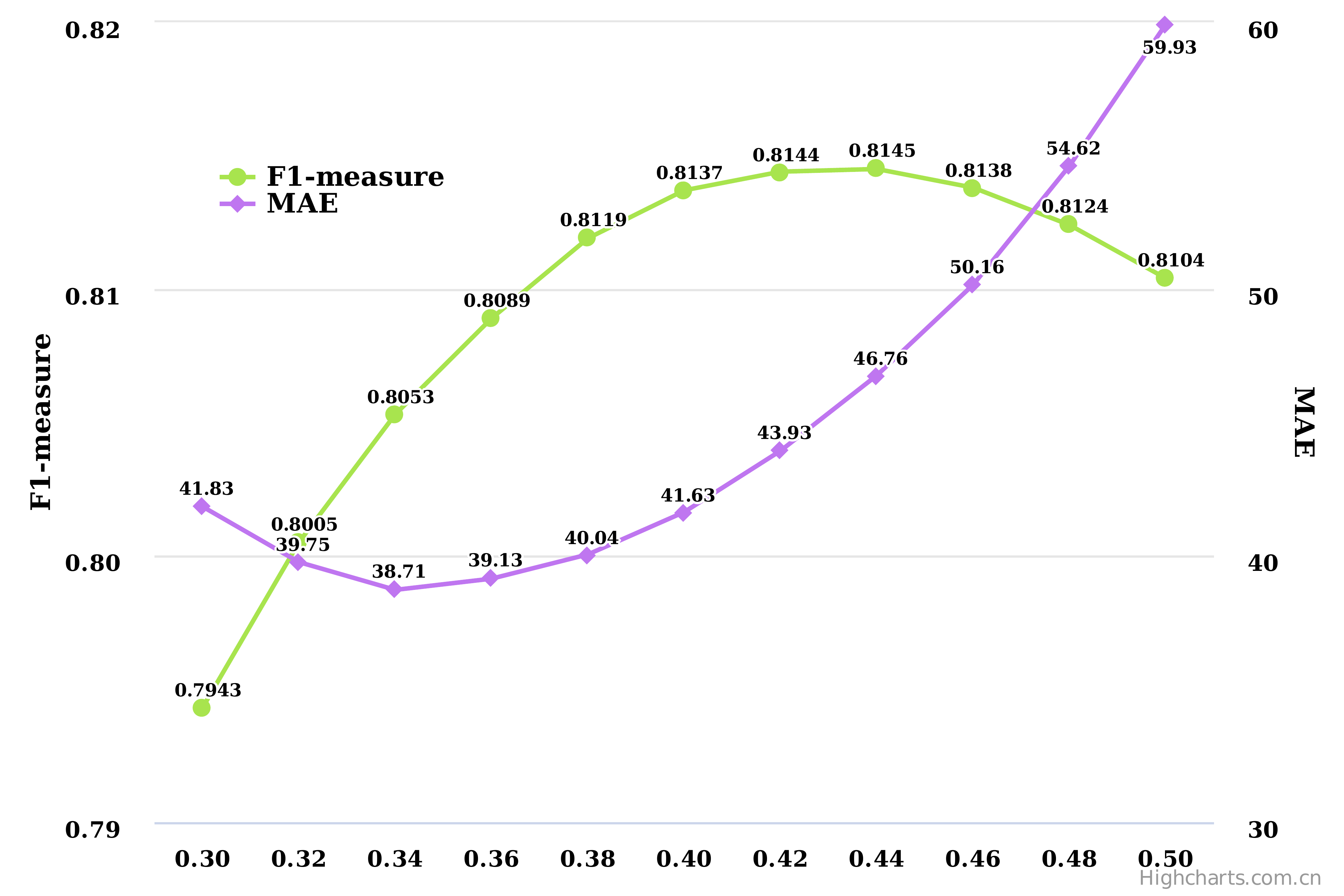}
	\caption{The performance of F1-measure and MAE under different threshold selections. }
	\label{fig:chart}
\end{figure}

\subsection{Trade-off between Localization and Counting Performance}

In the experiments, we find that the selection of threshold affects the performance for localization and counting significantly. For further exploring this problem, Fig. \ref{fig:chart} shows the F1-measure and the MAE by the way of line chart on the NWPU \emph{validation set}, where F1-measure and MAE represent the localization and counting performance, respectively. Notably, F1-measure and MAE is calculated under different threshold (in $\{ 0.30,\;0.32,\;0.34,\;...,\;0.50\}$). From the figure, we find the best F1-measure of 81.45\% and MAE of 38.71 locates at 0.44 and 0.32, respectively. 

The main reason is that there are more independent areas at low thresholds. For the localization task, the model tends to produce high-confidence prediction to achieve higher F1-measure, which results in missing many objects. Generally, the counting number is less than the GT in crowd scenes. As the threshold decreases, more independent regions are output by the model. Although many False Positives (FP) are included, the counting number is increased, which is closer to the GT in most cases. As a result, the best MAE is under a low-value threshold.

\subsection{Analysis of DCB's Location in Encoder}

Swin Transformer contains four stages, which have different dimensions. In this section, we analyze where Dilated Convolutional Block (DCB) should be added in the traditional Swin Transformer. To be specific, the four configurations are compared in Table \ref{Table:loc}: the checkmark means that DCB is added to the corresponding stage. All models are evaluated on the NWPU-Crowd \emph{validation set}. Each DCB consists of two dilated convolutional layers with the dilatation rate of 2 and 3, respectively.

From the table, when DCB is inserted into Stage 3 and 4, the performance is the best, namely F1-measure of 81.4, Precision of 84.1, and Recall of 79.0. Based on this setting, the performance is reduced when DCB is added to Stage 2 and 1. In fact, the shallow layer focus on learning local structure patterns. Introducing a large-range encoder (such as the proposed DCB) reduces the ability to extract local features of the model.

\subsection{Design of DCB}

\begin{table} [t]
	\centering
	\caption{The results (F1-measure/Pre./Rec.) of different Dilated Convolutional Backbones on the NWPU-Crowd \emph{validation set}.  (\%)}
	\begin{tabular}{ccccIccc}
		\whline
		Stage1	& Stage2	& Stage3	& Stage4	& F1-m & Pre. &Rec.\\
		\whline
		\rmark &\rmark &\rmark &\rmark & 79.2 & 81.6 & 76.9 \\	
		&\rmark &\rmark &\rmark & 80.0 &83.1 & 77.2 \\
		& &\rmark &\rmark & \textbf{81.4} & \textbf{84.1} & \textbf{79.0} \\
		& & &\rmark & 80.8 & 83.5 & 78.3 \\	
		\whline
	\end{tabular}
	\label{Table:loc}
\end{table}

The last section discusses the impact of DCB's location in Swin Transformer. Besides, the configuration inside DCB will affect the performance of the entire localization model. This section conducts the experiments of different dilatation rates in DCB on the NWPU \emph{validation set}. Specifically, following the HDC's idea \cite{wang2018understanding}, three groups of dilatation rate settings are designed: $\{ 2,\;2\} $, $\{ 2,\;3\} $ and $\{ 2,\;5\} $. The DCB is inserted into Stage 3 and 4 in Swin Transformer.

\begin{table} [h]
	\centering
	\caption{The results (F1-measure/Pre./Rec.) of different Dilated Convolutional Blocks on the NWPU-Crowd \emph{validation set}.  (\%)}
	\begin{tabular}{cIccc}
		\whline
		Dilatation Rate	& F1-m & Pre. &Rec.\\
		\whline
		$\{ 2,\;2\} $ & 80.1 & 87.5 & 73.8 \\	
		$\{ 2,\;3\} $ & \textbf{81.4} & \textbf{84.1} & \textbf{79.0} \\
		$\{ 2,\;5\} $ & 80.2 & 82.9 & 77.6 \\
		
		\whline
	\end{tabular}
	\label{Table:design}
\end{table}

Table \ref{Table:design} reports the localization results of the aforementioned settings. Notably, the dilatation rate of $\{ 2,\;3\} $ obtains the best performance, F1-measure of 81.4, Precision of 84.1, and Recall of 79.0. From the results, we find that too large or too small a receptive field will reduce the performance of the model. The former may learn more contextual information so that miss the local structure patterns, which will perform poorly in some clear scenes. The latter loses more large-range features, which causes that the model cannot handle dense and blurred crowd regions. Therefore, this paper selects the DCB with the dilatation rate of $\{ 2,\;3\} $ to achieve better performance.

\section{Conclusion}

\label{conclusion}

This paper proposes a joint of transformer and traditional convolutional networks method to tackle dense prediction problem for crowd localization. Notably, in Swin Transformer backbone, two dilated convolutional blocks are inserted in different stages to enlarge the respective field, which effectively prompts the capacity of feature extraction, especially for tiny objects, mutual occlusion, and blurred regions in the crowd scenes. Extensive experiments show the effectiveness of the proposed mechanism and achieve state-of-the-art performance on six mainstream datasets. Besides, this paper further discusses the relationship of performance between localization and counting tasks by some interesting experimental phenomena. In the future, we will focus on exploring the difference in learning parameters between the two tasks.


\bibliographystyle{IEEEtran}
\bibliography{IEEEabrv,reference}

\end{document}